\documentclass[10pt,twocolumn,letterpaper]{article}
\usepackage[pagenumbers]{cvpr} 

\usepackage{graphicx}
\usepackage{amsmath}
\usepackage{amssymb}
\usepackage{booktabs}
\usepackage{graphicx}
\usepackage{stfloats}
\usepackage{rotating}
\usepackage{subcaption}
\usepackage{multirow}
\usepackage{rotating}
\usepackage{subcaption}
\usepackage[pagebackref,breaklinks,colorlinks]{hyperref}

\usepackage[capitalize]{cleveref}
\crefname{section}{Section }{Secs.}
\Crefname{section}{Section}{Sections}
\Crefname{table}{Table}{Tables}
\crefname{table}{Tab.}{Tabs.}

\def\cvprPaperID{*****} 
\def\confName{CVPR}
\def\confYear{2023}

\begin{document}

\title{PointVector: A Vector Representation In Point Cloud Analysis}

\author{Xin Deng\thanks{Co-first authors with equal contribution to refining the theory and experimental design}\footnotemark[1]
\hspace{0.6cm}
WenYu Zhang\footnotemark[1]
\hspace{0.6cm}
Qing Ding\thanks{Corresponding authors}\footnotemark[2]
\hspace{0.6cm}
XinMing Zhang\footnotemark[2]\\
University of Science and Technology of China\\
{\tt\small \{xin\_deng,\,wenyuz\}@mail.ustc.edu.cn,}\,
{\tt\small \{dingqing,\,xinming\}@ustc.edu.cn}
}
\maketitle

\begin{abstract}
In point cloud analysis, point-based methods have rapidly developed in recent years. These methods have recently focused on concise MLP structures, such as PointNeXt, which have demonstrated competitiveness with Convolutional and Transformer structures. However, standard MLPs are limited in their ability to extract local features effectively. To address this limitation, we propose a Vector-oriented Point Set Abstraction that can aggregate neighboring features through higher-dimensional vectors. To facilitate network optimization, we construct a transformation from scalar to vector using independent angles based on 3D vector rotations. Finally, we develop a PointVector model that follows the structure of PointNeXt. Our experimental results demonstrate that PointVector achieves state-of-the-art performance $\textbf{72.3\% mIOU}$ on the S3DIS Area 5 and $\textbf{78.4\% mIOU}$ on the S3DIS (6-fold cross-validation) with only $\textbf{58\%}$ model parameters of PointNeXt. We hope our work will help the exploration of concise and effective feature representations. The code will be released soon.
\end{abstract}
\section{Introduction}
\begin{figure}[htbp]
	\centering
	\begin{subfigure}{0.49\linewidth}
		\centering
		\includegraphics[width=0.9\linewidth]{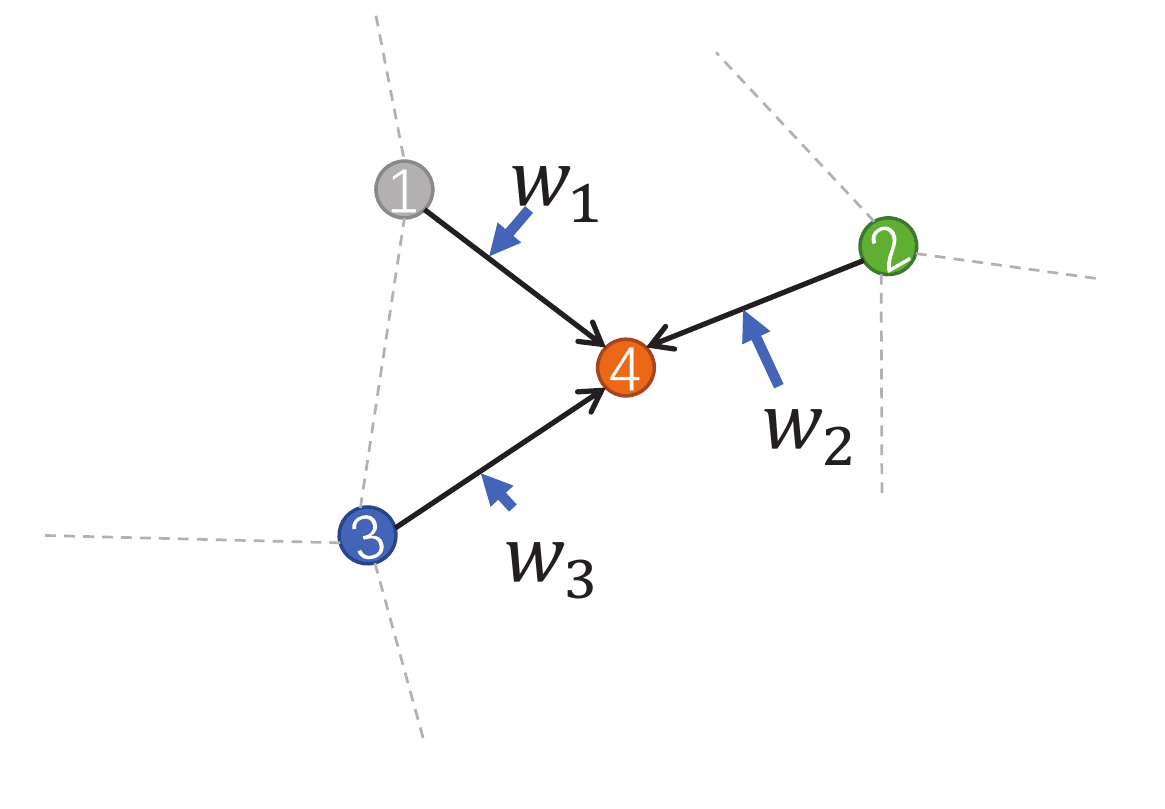}
		\caption{Attention}
		\label{fig:compare a}
	\end{subfigure}
	\centering
	\begin{subfigure}{0.49\linewidth}
		\centering
		\includegraphics[width=0.9\linewidth]{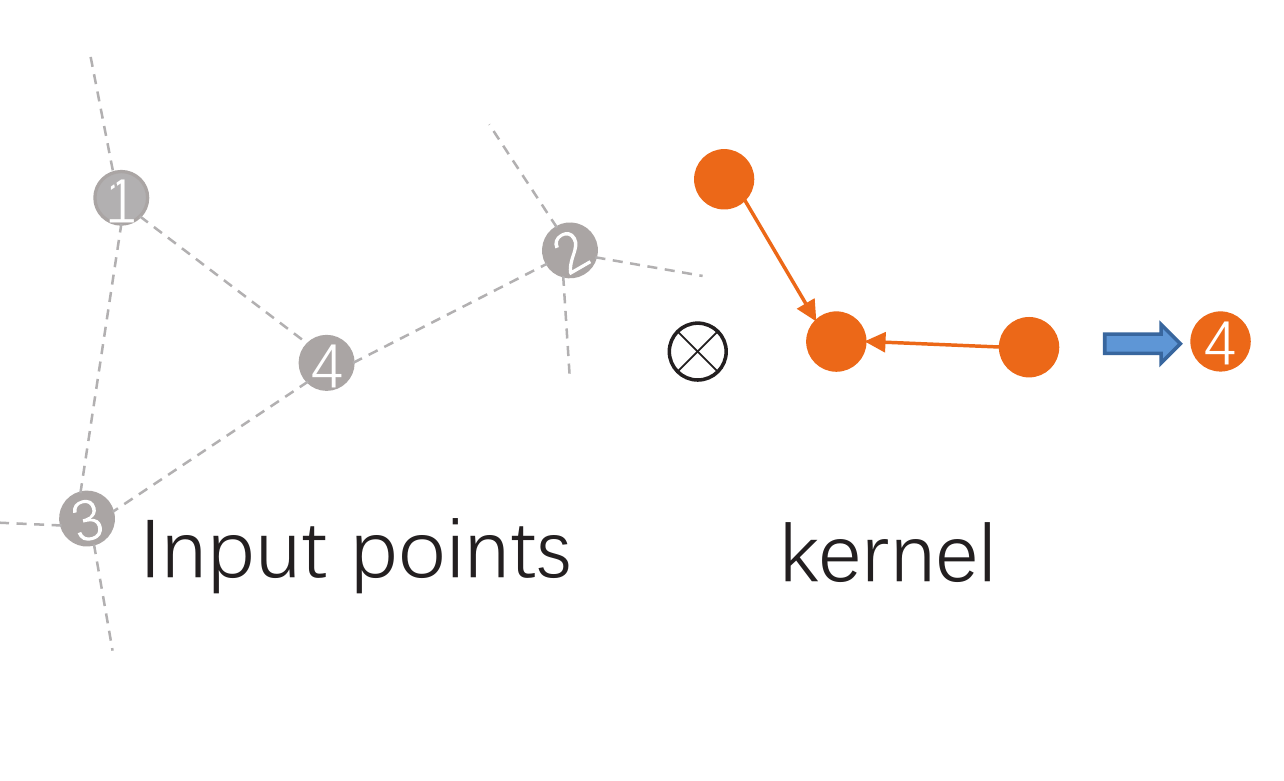}
		\caption{Templated-based method}
		\label{fig:compare b}
	\end{subfigure}
	\begin{subfigure}{0.49\linewidth}
		\centering
		\includegraphics[width=0.9\linewidth]{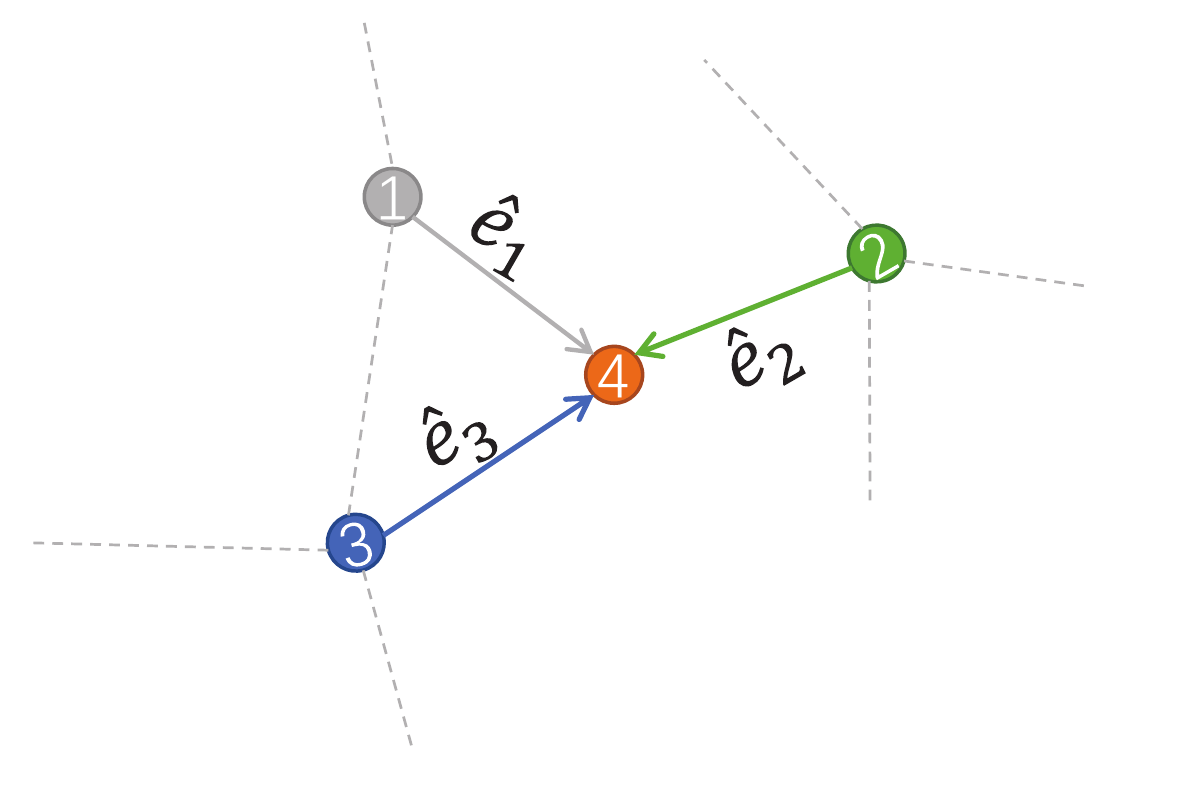}
		\caption{Dynamic Conv}
		\label{fig:compare c}
	\end{subfigure}
 \begin{subfigure}{0.49\linewidth}
		\centering
		\includegraphics[width=0.9\linewidth]{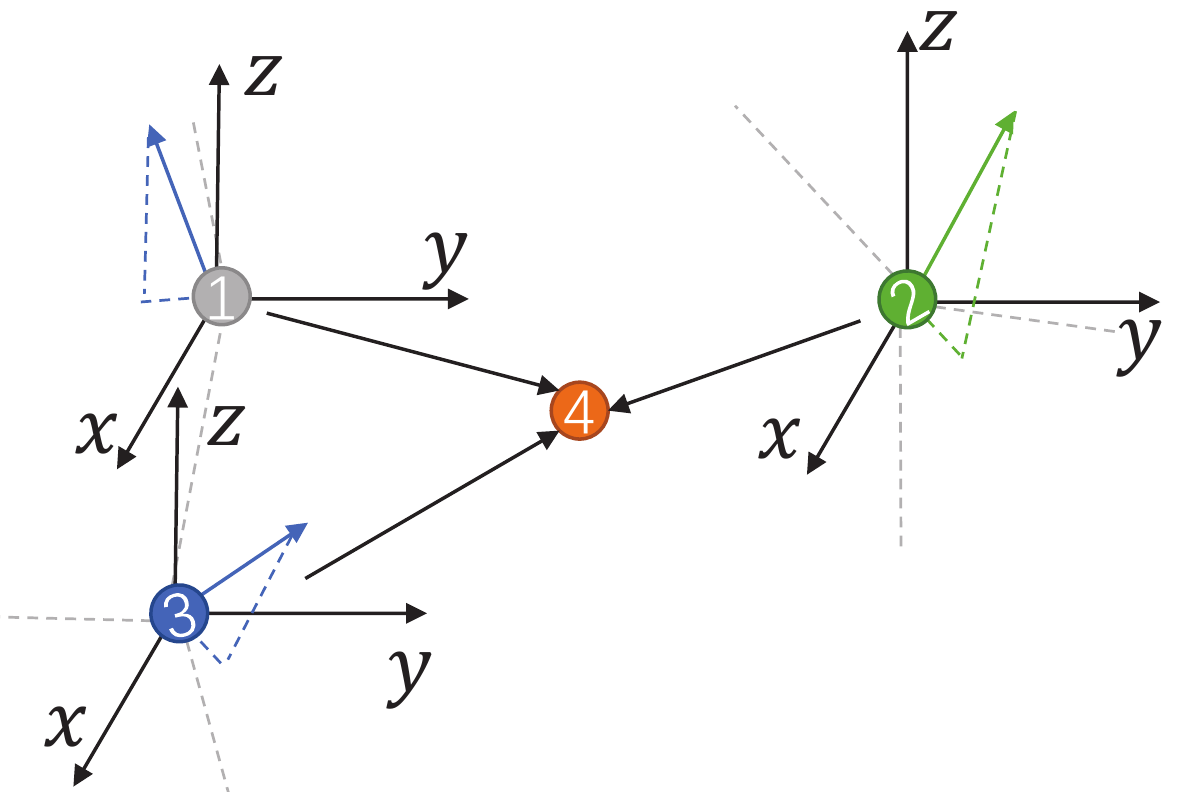}
		\caption{Vector}
		\label{fig:compare d}
	\end{subfigure}
	\caption{Illustrations of the core operations of the different methods. (a) The features of each point are calculated separately by applying a fixed/isotropic kernel (black arrow) like Linear layer. Then, it imparts anisotropy by weights generated from inputs. (b) The displacement vector is used to filter points that approximate the kernel pattern for features aggregation. (c) It applies unique dynamic kernels with anisotropy for each point feature. (d) Differently, we generate vector representations based on features, and the aggregation methods for vectors are anisotropic due to the direction of the vectors.}
	\label{fig:compare}
 \vspace{-5pt}
\end{figure}

Point cloud analysis is a cornerstone of various downstream tasks. With the introduction of PointNet\cite{2017.PointNet} and PointNet++\cite{2017.PointNet++}, the direct processing of unstructured point clouds has become a hot topic. Many point-based networks
introduced novel and sophisticated modules to extract local features, e.g., attention-based methods\cite{2021.PointTransformer} explore attention mechanisms as Fig.\ref{fig:compare a} with lower consumption, convolution-based methods\cite{2019.KPConv} explore the dynamic convolution kernel as Fig.\ref{fig:compare c},  and graph-based methods\cite{2019.DGCNN}\cite{2021.adaptconvpoint} use graph to model relationships of points. The application of these methods to the feature extraction module of PointNet++ brings an improvement in feature quality. However, they are somewhat complicated to design in terms of network structure. PointNeXt\cite{https://2022pointnext} adapts the SetAbstraction (SA) module of PointNet++\cite{2017.PointNet++} and proposes the Inverted Residual MLP (InvResMLP) module.
The simple design of MLP network achieves good results. Motivated by this work, we try to further explore the potential of the MLP structure.

PointNeXt uses all standard MLPs, which has insufficient feature extraction capability. In addition to attention and dynamic convolution mechanisms, template-based methods as Fig.\ref{fig:compare b} such as 3D-GCN\cite{3D-GCN} employ relative displacement vectors to modulate the association between input points and the convolutional kernel. 
We introduce a vector representation of features to extend the range of feature variation with the intention of more effectively regulating the connections between local features. Our approach as Fig.\ref{fig:compare d} differs from template-based methods. Instead of using displacement vectors as a property of the kernel, we generate a vector representation for each neighboring point and aggregate them. Our method introduces less inductive bias, resulting in improved generalization capabilities. Furthermore, we enhance the generation of 3D vector representations by utilizing a vector rotation matrix with two independent angles in 3D space. This method facilitates the network to find the better solution.

Influenced by PointNeXt\cite{https://2022pointnext} and PointNet++\cite{2017.PointNet++}, we present the VPSA module. This module adheres to the structure of Point Set Abstraction (SA) module of the PointNet series. Vector representations are obtained from input features and aggregated using a reduction function. The vector of each channel is then projected into a scalar to derive local features. By combining VPSA and SA modules, we construct a PointVector model with an architecture akin to that of PointNeXt.

Our model undergoes comprehensive validation on public benchmark datasets. It achieves state-of-the-art performance on the S3DIS\cite{s3dis} semantic segmentation benchmark and competitive results on the ScanObjectNN\cite{2016.Scan} and ShapeNetPart\cite{2016.shapenet} datasets. By incorporating a priori knowledge of vectors, our model attains superior results with fewer parameters on S3DIS. Detailed ablation experiments further demonstrate the efficacy of our methodology. The contributions are summarized below:
\begin{itemize}
    \item [-] We propose a novel immediate vector representation with relative features and positions to better guide local feature aggregation.
    \item[-] We explore the method of obtaining vector representation and propose the generation 
method of 3D vector by utilizing the vector rotation matrix in 3D space.
    \item[-] Our proposed PointVector model achieves $\textbf{72.3\%}$ mean Intersection over Union (mIOU) on S3DIS area5 and $\textbf{78.4\%}$ mIOU on S3DIS (6-fold cross-validation) with only $\textbf{58\%}$ model parameters of PointNeXt.
\end{itemize}
\section{Related work}
\paragraph{Point-based network. }In contrast to the voxelization\cite{2018.VoxelNet}\cite{2019.PointPillars}\cite{2020.PV-RCNN} and multiview\cite{2015.MVCNN}\cite{2021.MVTN}\cite{2021.IMVCNN} methods, point-based methods deal directly with point clouds. PointNet first proposes using MLP to process point clouds directly. PointNet++ subsequently introduces a hierarchical structure to improve the feature extraction. Subsequent works focused on the design of fine-grained local feature extractors. Graph-based methods\cite{2019.DGCNN}\cite{2019.GAC} rely on a graph neural network and introduce point features and edge features to model local relationships. Conv-based methods\cite{2019.KPConv}\cite{2018.SpriderCNN}\cite{2019.PointConv}\cite{2018.PCNN}\cite{2018.PointCNN} propose several dynamic convolution kernels to adaptively aggregate neighborhood features. Many transformer-like networks\cite{2020.SVGA-Net}\cite{2021.PointBERT}\cite{2021.PoinTr}\cite{2021.PCT}\cite{2022.straitifiedformer} extract local features with self-attention. Recently, MLP-like networks are able to obtain good results with simple networks by enhancing the features. PointMLP\cite{2022.PointMLP} proposes a geometric affine module to normalize the feature. RepSurf\cite{repsurf} fits the surface information through the triangular plane, models umbrella surfaces to provide geometric information. PointNeXt\cite{https://2022pointnext} integrates training strategies and model scaling.
\paragraph{MLP-like Architecture.}The MLP-like structure has recently shown the ability to rival the Transformer with simple architecture. In the image field, MLP-Mixer\cite{2021MLP-mixer} first use the combination of Spatial MLP and Channel MLP. The subsequent works\cite{chen2022cyclemlp}\cite{Lian_2021_ASMLP} reduce computational complexity by selecting objects for the spatial MLP while maintaining a large perceptual field to preserve accuracy. Since the point cloud is too large, the MLP-like network determines the perceptual field generally using K-Nearest neighbor sampling or ball sampling methods. The MLP structure in point cloud analysis starts with PointNet\cite{2017.PointNet} and PointNet++\cite{2017.PointNet++}, using MLPs to extract features and aggregating them by symmetric functions. Point-Mixer\cite{2021.PointMixer} proposes three point-set operators, PointMLP\cite{2022.PointMLP} to modify the distribution of features by geometric affine module, and PointNeXt\cite{https://2022pointnext} to scale up the PointNet++ model and improve the performance using by training strategies and model scaling.
\paragraph{Feature Aggregation. }
PosPool\cite{2020.pospool} improves the reduction function defined in PointNet++ by providing a parameter-free position-adaptive pooling operation. ASSANet\cite{assanet} introduces a new anisotropic reduction function. Also, the introduction of the attention mechanism\cite{PointAtt} provides new dynamic weights for the reduction function. Vectors have direction, and this property is naturally satisfied for anisotropic aggregation functions. GeoCNN\cite{geocnn} projects features based on vectors and angles of neighbor points and centroids in six directions and sums them. WaveMLP\cite{2022WaveMLP} represents image patches as waves and describes feature aggregation using wave phase and amplitude. The Vector Neuron\cite{vectorneurons} constructs a triad of neurons to reconstruct standard neural networks and represent features through vector transformations. The template-based methods represented by 3DGCN\cite{3D-GCN} uses the cosine value of the relative displacement vectors to filter for aggregation features from neighbors that more conform to the pattern of the kernel. Local displacements\cite{wang2022displacement} use local displacement vectors to update features by combining the weights of fixed kernels. In our method, an intermediate vector representation is generated by modifying the point feature extraction function. The vector direction is determined based on both features and position to fulfill the anisotropic aggregation function.
\section{Method}
We propose an intermediate vector representation to enhance local feature aggregation in point cloud analysis. This section includes a review of the Point Set Abstraction(SA) operator of the PointNet family in Section \ref{preliminary}, the presentation of our Vector-oriented Point Set Abstraction module in Section \ref{vpsa}, a description of our method of extending vectors from scalars in Section \ref{extend vector}, and the network structure of PointVector in Section \ref{architecture}.
\subsection{Preliminary}\label{preliminary}
The SA module include a grouping layer (K-NN or Ball-Query) to query each point’s neighbors, shared MLPs, and a reduction layer to aggregate neighbor features. The SA module has an subsample layer to downsample the point cloud in the first layer. We denote $f_i^{l+1}$ as the extracted feature of point \emph{i} after stage \emph{l+1}, $N_i$ as the neighbors of point \emph{i} and $n$ is the number of incoming points. The content of the SA module can be formulated as follows:
\begin{equation}\label{eq:sa module}
    f_i^{l+1}=R\{H\{[f_j^l,p_j-p_i]\}|j\in N_i\},
\end{equation}
where $R$ is the reduction function that aggregates features for point \emph{i} from its neighbors $N_i$ and $H$ means the shared MLPs. $f_j^l,p_j,p_i$ denote the input features of point \emph{j}, the position of point \emph{j} and the position of point \emph{i}, respectively.

In the local aggregation operation, the classical method assigns weights to components of c-dimensional features as shown in Eq.\ref{1a} and sums the neighboring features in spatial dimensions. 
We consider the component $f_i$ of the \emph{c}-dimensional feature \emph{f} as a base vector with only one non-zero value, and define the vector transformation as follows:
\setlength{\arraycolsep}{1.5pt}
\setlength{\abovedisplayskip}{1pt}
\setlength{\belowdisplayskip}{1pt}
\begin{align}
f_i*w&=wf_i, i=0\cdots c,\label{1a}\\
    \begin{bmatrix}
    f_i&0&\cdots&0
    \end{bmatrix}
    \begin{bmatrix}
        w&0&\cdots&0\\
        \vdots&\vdots&\ddots&\vdots\\
        0&0&\cdots&0
    \end{bmatrix}
    &=
    \begin{bmatrix}
        wf_i&0&\cdots &0
    \end{bmatrix},\label{1b}
    \end{align}
where \emph{w} is the scalar weight. In Eq.\ref{1b}, the transformation changes one value of the vector. The two equations above are equivalent. Unchanged zeros in the equation do not contribute to subsequent operations and can be disregarded. \emph{In Physics, the degree of freedom of a motion is equal to the number of state quantities that the motion causes the system to change. }A greater number of degrees of freedom in a physical system indicates a larger range of independent variation in the parameters that define its state. Similarly, the degrees of freedom of a vector transformation refer to the number of values in the vector that can change independently. \textbf{So, the 3D vector we mentioned means the degrees of freedom of the vector transformation is 3.}
\subsection{Vector-oriented Point Set Abstraction}\label{vpsa}
\begin{figure}[htbp]
    \centering
    \includegraphics[width=\linewidth]{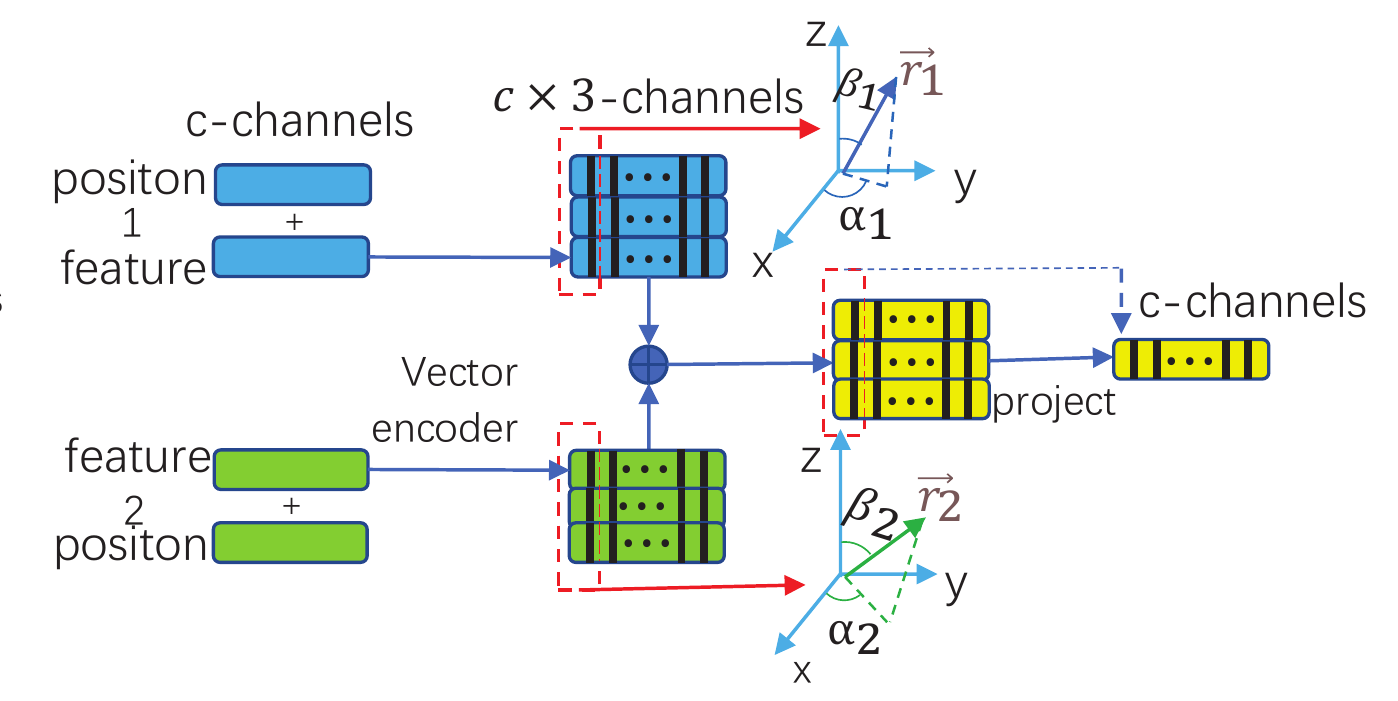}
    \caption{The vector-oriented point set abstraction (VPSA) module of PointVector. 
    It illustrates that VPSA module obtains vector representations from input features, aggregates them, and projects them back to the original feature style. As shown in the figure, each channel of the feature can be considered a 3D vector, with channels being independent of one another.
    }
    \label{fig:vector and point}
    \vspace{-0.3cm}
\end{figure}
As discussed in Section \ref{preliminary}, feature components can be represented as vectors. A higher degree of freedom in vector transformations allows for increased variation and improved representation of connections between neighboring elements. 
 Vectors, with their size and direction properties, are more expressive than scalars for representing features. When aggregated, they exhibit anisotropy due to their directional nature. So, we introduce an intermediate vector representation as Fig.\ref{fig:vector and point}. 
 
 It should be noted that in our assumptions, the component of a \emph{c}-dimensional feature represents the projection of the feature vector along the \emph{c} coordinate axes. After aggregating the vectors to obtain the $c\times3$ centroid feature, where the number of changing values in the component vectors is 3. To merge them into a \emph{c}-dimensional feature vector requires aligning the c components and then summing them. Due to the difficulty in implementing component alignment with this method, we directly project the \emph{c} components into scalars and combine them into centroid features. Similar to the intermediate features in a convolutional network, the values on each channel's feature map represent the response strength to a specific feature at that location.

The input features in our method are transformed into a series of vectors and then aggregated by the reduction function. Note that the element in each channel of the vector representation is vector. We obtain a vector representation that is channel independent. We denote ${fp}_j$ as a mixed feature of relative features $f_j-f_i$ and relative positions $p_j-p_i$. The content of the vector-guided aggregation module can be formulated as:
\begin{equation}\label{eq:vector}
    f_i^{l+1}=\eta(f_i^l)+H_c\{H_p\{R\{H_v(fp_j)|j\in N_i\}\}\},
\end{equation}
where $H_v$ is the function that generates the vector representation, $H_p$ denotes the projection Linear transform vector to a scalar, and $H_c$ is the channel mixing Linear that interacts with the information of each channel while transforming dimensions to fit the network. However, the feature representation we introduce is actually represented using a triplet form. We denote \emph{m} as the dimension of the vector, and $c$ is the channel of the feature. In fact, the set of \emph{c} \emph{m}-dimensional vectors is represented in the same form as the $(m\times c)$-dimensional feature vectors. The reduction function is followed by a grouped convolution\cite{Groupconv.imagenet.deepcnn} that transforms the vectors to scalars for each channel, which distinguishes the intermediate vector representation from the general feature vector.

When the reduction function $R$ selects sum, the $R$ and $H_p$ functions together constitute a special case of GroupConv \cite{Groupconv.imagenet.deepcnn}. Let $k$ denote the number of neighbor features. For one group, the convolution kernel of GroupConv is a $k\times m\times1$ parameter matrix, while our method can be viewed as \emph{k} identical $m\times 1$ parameter matrices. This is because we treat vectors as wholes and assign equal weight to each element. 
We will explain in the supplementary material why the original groupconv operation is not suitable for our vector-guided feature aggregation.

\subsection{Extended Vector From Scalar}\label{extend vector}
The simplest idea for the $H_v$ function defined in Eq.\ref{eq:vector} is to obtain \emph{c} \emph{m}-dimensional vectors of point \emph{j} directly with MLPs. However, while single-layer MLPs may have limited expressive capability, multi-layer MLPs can be resource-intensive. As discussed in Section \ref{preliminary}, input features are considered as vectors and we aim to design a transformation with high degrees of freedom. This transformation combines rotation and scaling, represented by a rotation matrix and a learnable parameter respectively. This method achieves better results with lower resource consumption.
\begin{figure}
    \centering
    \includegraphics[width=\linewidth]{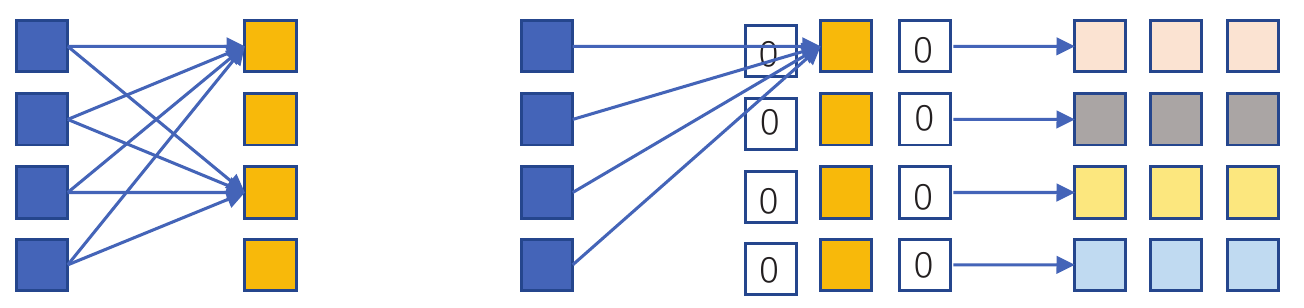}
    \caption{Extension from general feature to vector representation. For simplicity, we tentatively set c=4 and m=3. The left side represents the process of generating features by standard MLP, and the right side adds 2 components to each scalar of the features to form a vector and then rotates it.}
    \label{fig:extend}
    \vspace{-0.3cm}
\end{figure}

As shown in Fig.\ref{fig:extend}, a scalar can be directly converted into an m-dimensional vector by adding m-1 zero-value components. Each channel of the extended vector representation can then be considered as an m-dimensional vector along a specific coordinate axis direction. 
Therefore, we can obtain the proper vector direction by additionally training a rotation matrix.
\begin{figure}[htbp]
    \centering
    \includegraphics[width=0.7\linewidth]{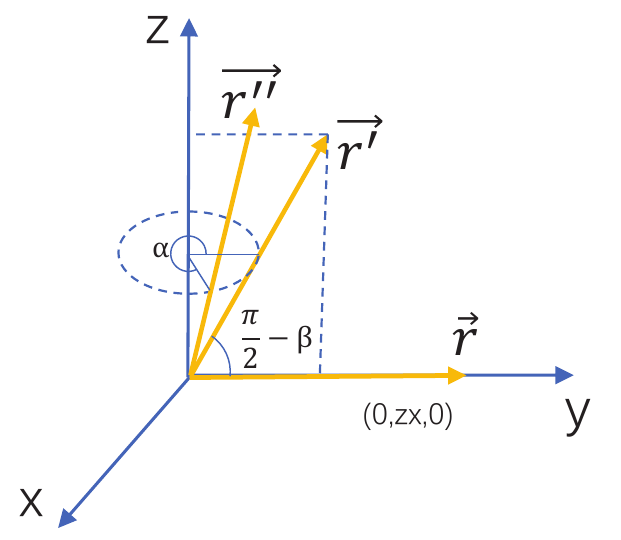}
    \caption{The rotation of a 3D vector. The vector $\Vec{r}$ can be obtained by two rotations to obtain another vector $\vec{r''}$}
    \label{fig:rotation}
    \vspace{-0.3cm}
\end{figure}
Directly predicting the rotation matrix can cause difficulties for nonlinear optimization because the matrix elements are interdependent. Instead, we first predict the rotation angle and then derive the rotation matrix based on this angle. The rotation of a 3D vector can be decomposed into rotations around three axes. However, we have not yet determined how to represent the rotation of a 4D vector around a plane. As shown in Fig.\ref{fig:rotation}, since the extended 3D vector is on the coordinate axis, one rotation around that axis can be omitted. We keep the default rotation direction as counterclockwise. The vector $\Vec{r}$ is first rotated around the \emph{x}-axis by an angle $\pi/2 - \beta$ and then rotated around the \emph{z}-axis by an angle $\alpha$ to finally obtain the vector $\overrightarrow{r''}$. The rotation can be formulated as follows:
\begin{equation}
\small
\begin{aligned}\label{eq:transform}
    \overrightarrow{r''}&={Rot}_z{Rot}_x\Vec{r}\\
    &=
    \begin{bmatrix}
    \begin{smallmatrix}
    cos(\alpha) & -sin(\alpha) & 0\\
    sin(\alpha)&cos(\alpha)&0\\
    0&0&1
    \end{smallmatrix}
    \end{bmatrix}
    \begin{bmatrix}
    \begin{smallmatrix}
    1&0&0\\
    0&sin(\beta) & -cos(\beta)\\
    0&cos(\beta)&sin(\beta)\\
    \end{smallmatrix}
    \end{bmatrix}
    \begin{bmatrix}
    \begin{smallmatrix}
    0\\
    zx\\
    0
    \end{smallmatrix}
    \end{bmatrix}\\
    &=
    \begin{bmatrix}
    \begin{smallmatrix}
    cos(\alpha) & -sin(\alpha)sin(\beta) &sin(\alpha)cos(\beta)\\
    sin(\alpha)&cos(\alpha)sin(\beta)&-cos(\alpha)cos(\beta)\\
    0&cos(\beta)&sin(\beta)
    \end{smallmatrix}
    \end{bmatrix}
    \begin{bmatrix}
    \begin{smallmatrix}
    0\\
    zx\\
    0
    \end{smallmatrix}
    \end{bmatrix}\\
    &=
    \begin{bmatrix}
    \begin{smallmatrix}
    -zx\cdot sin(\alpha)sin(\beta)\\
    zx\cdot cos(\alpha)sin(\beta)\\
    zx \cdot cos(\beta)
    \end{smallmatrix}
    \end{bmatrix},
\end{aligned}
\end{equation}
where $Rot_x,Rot_z$ denote the rotation matrix rotated around the x-axis and the rotation matrix rotated around the z-axis, respectively, and $zx$ is generated by Linear. The independence of $\alpha$ and $\beta$ facilitates network optimization.

Therefore, we can expand each scalar value of the features into a 3D vector according to Fig.\ref{fig:rotation} and Eq.\ref{eq:transform}.
The feature aggregation in a local area is influenced by the relationship between neighboring points and centroids. Methods such as PointTransformer\cite{2021.PointTransformer}, PAConv\cite{2021.PAConv}, and Adaptconv\cite{2021.adaptconvpoint} model this relationship using relative position and features. Our approach also extracts rotation angles using MLP on relative positions and features. The acquisition of the vector can be formulated as follows:
\begin{equation}
    \begin{aligned}\label{eq:zx_alpha_beta}
    zx_j&=Linear(fp_j)\\
    [\alpha_j,\beta_j]&=Relu(BN(Linear([fp_j]))),
    \end{aligned}
\end{equation}
where $fp_j$ denotes a mixed feature of relative features $f_j-f_i$ and relative positions $p_j-p_i$, and $f_j$ means the feature of point \emph{j}. Therefore, we can obtain the intermediate vector representation from the input features and positions by using Eq.\ref{eq:transform} and Eq.\ref{eq:zx_alpha_beta}.
\subsection{Architecture}\label{architecture}
\begin{figure*}[ht]
    \centering
    \includegraphics[width=0.9\linewidth]{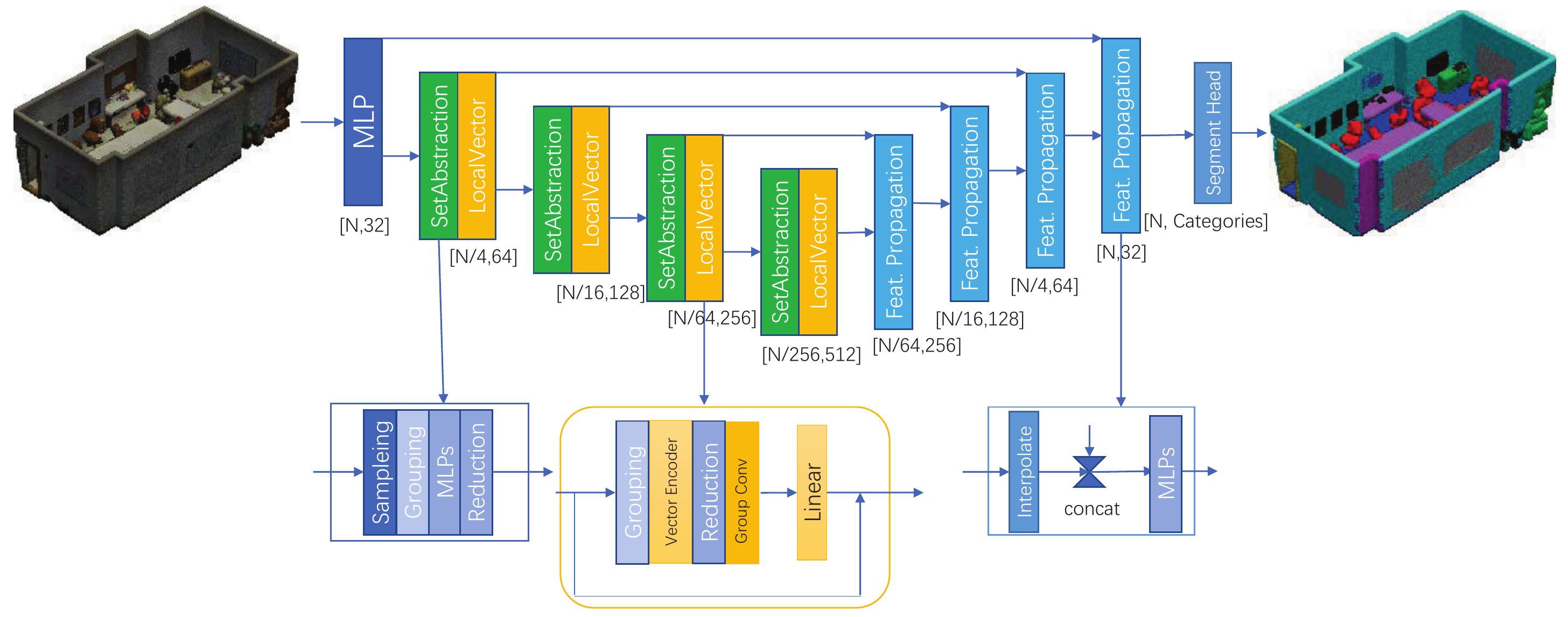}
    \caption{Overall Architecture. We reuse the SA module and Feature Propagation module of PointNet++ and propose the VPSA module to improve the feature extraction of sampled point clouds.}
    \label{fig:architecture}
    \vspace{-5pt}
\end{figure*}
In summary, we propose PointVector, modified from PointNeXt\cite{https://2022pointnext} by replacing its InvResMLP module with our proposed VPSA module, we defining its vector dimension $m=3$. The architecture is illustrated in Fig.\ref{fig:architecture}. Referring to the classical PointNet++, we use a hierarchical structure containing an encoder and a decoder. For the segmentation task, we use an encoder and a decoder. For the classification task, we only use an encoder. For a fair comparison with PointNeXt, we set up three sizes of models with reference to the parameter settings of PointNeXt. We denote C as the channel of embedding MLP in the beginning, S as the numbers of the SA module, V as the numbers of the VPSA module. The three sizes of models are shown as follows:
\begin{itemize}
    \item PointVector-S: C=32, S=0, V=[1,1,1,1]
    \item PointVector-L: C=32, S=[1,1,1,1], V=[2,4,2,2]
    \item PointVector-XL: C=64, S=[1,1,1,1], V=[3,6,3,3]
\end{itemize}
Since PointNeXt uses only the PointNeXt-S model for classification, we use our VPSA module instead of the SA module in PointVector-S for a fair comparison. The detailed structure of the classification tasks will appear in the supplementary material. There is a skip connection path in the VPSA module in Fig.\ref{fig:architecture}, which is added to the main path and then through a ReLU layer. The reason for using this summation method is that RepSurf\cite{repsurf} indicates how two features with different distributions should be combined. For the segmentation task, finer local information is needed, and we set reduction function as sum. For the classification task, which favors aggregating global information, we choose the original reduction function such as max. 

\section{Experiments}
We evaluate our model on three standard benchmarks: S3DIS\cite{s3dis} for semantic segmentation and ScanObjectNN\cite{2016.Scan} for real-world object classification and ShapeNetPart\cite{2016.shapenet} for part segmentation. Note that our model is implemented on the basis of PointNeXt. Since we use the training strategy provided by PointNeXt, we refer to the metrics reported by PointNeXt for a fair comparison.
\paragraph{Experimental setups.} We train PointVector using CrossEntropy loss with label smoothing\cite{label.smoothing}, AdamW optimizer\cite{adamw}, and initial learning rate lr=0.002, weight\_decay $10^{-4}$, with Cosine Decay, and a batch\_size of 32. The above are the base settings for all tasks, and specific parameters will be changed for specific tasks. We follow the train, valid, and test divisions for the dataset. The best model on the validation set will be evaluated on the test set. For S3DIS segmentation task, point clouds are downsampled with a voxel size of 0.4 m following previous methods\cite{2019.KPConv}\cite{assanet}\cite{2021.PointTransformer}. The initial learning rate on this task is set to 0.01. For 100 epochs, we use a fixed 24000 points as a batch and set batch\_size to 8. During training, the input points are selected from the nearest neighbors of the random points. Similar to Point Transformer\cite{2021.PointTransformer}, we evaluate our model using the entire scene as input. For ScanObjectNN\cite{2016.Scan} classification task, we set the weight\_decay to 0.05 for 250 epochs. Following Point-BERT\cite{2021.PointBERT}, the number of input points is 1024. The training points are randomly sampled from the point cloud, and the testing points are uniformly sampled during evaluation. The details of data augmentation are the same as those in PointNeXt. For ShapeNetPart part segmentation, we train PointVector-S with a batch size of 32 for 300 epochs. Following PointNet++\cite{2017.PointNet++}, $2,048$ randomly sampled points with normals are used as input for training and testing. 

For voting strategy\cite{2019.RS-CNN}, we keep it the same as PointNeXt and use it only on part segmentation task. To ensure a fair comparison with standard methods, we do not use any ensemble methods, such as SimpleView\cite{2021.Simple-View}. We also provide the model parameters (Params) and GFLOPs. We additionally, similar to PointNeXt, provide throughput (instance per second) as an indicator of inference speed. The input data for the throughput calculation are kept consistent with PointNeXt for fair comparison. The throughput of all methods is measured using 128 × 1024 (batch size 128, number
of points 1024) as input on ScanObjectNN and 64 × 2048 on ShapeNetPart. On S3DIS, 16 × 15,000 points are used to measure the throughput following \cite{https://2022pointnext}\cite{assanet}. We evaluate our model using an NVIDIA Tesla V100 32 GB GPU and a 48 core Intel Xeon @ 2.10 Hz CPU. 

\subsection{3D Semantic segmentation on S3DIS}
S3DIS\cite{s3dis} (Stanford Large-Scale 3D Indoor Spaces) is a challenging benchmark composed of 6 large-scale indoor areas, 271 rooms, and 13 semantic categories in total. For our models in S3DIS, the number of neighbors in SetAbstraction is 32, and the number of neighbors in the Local Vector module is 8. PointTransformer\cite{2021.PointTransformer} also employs most of the training strategies and data enhancements used by PointNeXt, so it is fair for us to compare with it. For a comprehensive comparison, we report the experimental results of PointVector-L and PointVector-XL on S3DIS with 6-fold cross-validation in Table \ref{tab:6-fold} and S3DIS Area 5 in Table \ref{tab:area5}, respectively. As shown in table \ref{tab:6-fold}\&\ref{tab:area5}, we achieve state-of-the-art performance on both validation options. Table \ref{tab:6-fold} shows that our largest mode PointVector-XL outperforms PointNeXt-XL by \textbf{1.6\%, 3.1\% and 3.5\% in terms of overall accuracy (OA), mean accuracy(mAcc) and mIOU, respectively, while has only 58\% Params}. At the same time, the computational consumption of ours is only \textbf{69\%} of PointNeXt-XL in terms of GFLOPs. The reduction in computational consumption because the number of neighbors is reduced to 8. \emph{The limitation is that we make heavy use of GroupConv (groups=channel), which is not well optimized in PyTorch and is slower than standard convolution.} Therefore, \textbf{our inference speed is 6 instances/second lower} than PointNeXt-XL. Our model shows better results at all sizes.

\begin{table}[htbp]
\small
    \centering
    \resizebox{\linewidth}{!}{
    \begin{tabular}{l|ccc|ccc}
    \hline
         \textbf{Method}& OA&mAcc& mIOU& Params&FLOPs&Throughput \\
         & \%&\%&\%&M&G&(ins./sec.)\\
         \hline
         PointNet\cite{2017.PointNet}&78.5&66.2&47.6&3.6&35.5&162\\
         PointCNN\cite{2018.PointCNN}& 88.1 & 75.6& 65.4   & 0.6 & - & -\\
         DGCNN\cite{2019.DGCNN}& 84.1 & - &56.1  & 1.3 & - & 8 \\
         DeepGCN\cite{deepgcn}  & 85.9 & -&60.0& 3.6 & - & 3 \\
         KPConv\cite{2019.KPConv} & - & 79.1 &70.6 & 15.0 & - & 30  \\
         RandLA-Net\cite{randlla_net} & 88.0 & 82.0&70.0  & 1.3  & 5.8 & 159 \\
         Point Transformer\cite{2021.PointTransformer} & 90.2 & 81.9 &73.5& 7.8 & 5.6  & 34 \\ 
         CBL\cite{CBL} & 89.6 & 79.4&73.1 & 18.6  &- & -\\
         RepSurf\cite{repsurf}&90.9 & 82.6&74.3 & 0.976 & -&-\\
         \hline
         PointNet++\cite{2017.PointNet++} & 81.0 & 67.1& 54.5 & 1.0 & 7.2 & 186 \\
         PointNeXt-L\cite{https://2022pointnext}&89.8&82.2&73.9&7.1 & 15.2  & 115 \\
         PointNeXt-XL\cite{https://2022pointnext}  & 90.3 & 83.0&74.9& 41.6 & 84.8 & 46 \\
         \textbf{PointVector-L}&\textbf{91.4}&\textbf{85.5}&\textbf{77.4}&4.2&10.7&98\\
         \textbf{PointVector-XL(Ours)}&\textbf{91.9}&\textbf{86.1}&\textbf{78.4}&24.1&58.5&40\\
         \hline
    \end{tabular}}
    \caption{\textbf{Semantic segmentation on S3DIS with 6-fold cross-validation.} Methods are in chronological order. The highest and second scores are marked in bold.}
    \label{tab:6-fold}
    \vspace{-0.3cm}
\end{table}
On S3DIS Area 5, we selected the best results reported by PointNeXt for comparison and did not repeat the experiment. Our PointVector-XL model outperforms StratifiedFormer\cite{2022.straitifiedformer} and PointNeXt-XL by \textbf{0.3\% and 1.8\%} in mIOU, respectively. StratifiedFormer expands the scope of the query by combining high-resolution and low-resolution keys while efficiently extracting contextual information. Even though its \emph{receptive field is much wilder} than our model, we still show a competitive performance. Additionally, there are some differences in the experimental setup between our model and it, in which it has \textbf{80k points} of input, much larger than our  \textbf{24k points} of input. In addition it uses \emph{KPConv\cite{2019.KPConv} instead of Linear} in the first layer. It seems that these measures have significant effects. However, the comparison is not fair enough for us due to the difference of the experimental configurations. We will synchronize its experimental configuration later. Additionally, our models of the same size on Area 5 show better results than PointNeXt. PointVector-L and PointVector-XL perform better than PointNeXt-L and PointNeXt-XL by 1.7\% and 1.5\% in mIOU, respectively, and we performs better on most of categories.
\begin{table*}[ht]
\small
    \centering
    \resizebox{0.9\linewidth}{!}{
    \begin{tabular}{ l | c c c | c c c c c c c c c c c c c}
    \hline
         \textbf{Method}
&\begin{sideways}   OA   \end{sideways}
&\begin{sideways}   mAcc \end{sideways}
&\begin{sideways}   mIoU \end{sideways}
&\begin{sideways}   ceiling \end{sideways}
&\begin{sideways}   floor \end{sideways}
&\begin{sideways}   wall \end{sideways}
&\begin{sideways}   beam \end{sideways}
&\begin{sideways}   column\end{sideways}
&\begin{sideways}   window \end{sideways}
&\begin{sideways}   door \end{sideways}
&\begin{sideways}   table \end{sideways}
&\begin{sideways}   chair \end{sideways}
&\begin{sideways}   sofa \end{sideways}
&\begin{sideways}   bookcase \end{sideways}
&\begin{sideways}   board \end{sideways}
&\begin{sideways}   clutter\end{sideways} \\
         & \%&\%&\%\\
         \hline
         PointNet\cite{2017.PointNet}& - & 49.0 & 41.1 & 88.8 & 97.3 & 69.8 & 0.1 & 3.9 & 46.3 & 10.8 & 59.0 & 52.6 & 5.9 & 40.3 & 26.4 & 33.2\\
         PointCNN\cite{2018.PointCNN}& 85.9 & 63.9 & 57.3  & 92.3 & 98.2 & 79.4 & 0.0 & 17.6 & 22.8 & 62.1 & 74.4 & 80.6 & 31.7 & 66.7 & 62.1 & 56.7\\
         DGCNN\cite{2019.DGCNN}& 83.6 & - & 47.9 & - & - & - & - & - & - & - & - & - & - & - & - & - \\ 
         DeepGCN\cite{deepgcn}& - & - & 52.5 & - & - & - & - & - & - & - & - & - & - & - & - & -   \\ 
         KPConv\cite{2019.KPConv}& - & 72.8 & 67.1 &92.8 & 97.3 & 82.4 & 0.0 & 23.9 & 58.0 & 69.0 & 81.5 & 91.0 & 75.4 & 75.3 & 66.7 & 58.9 \\ 
         PVCNN\cite{2019.pvcnn}& 87.1  & - & 59.0 & - & - & - & - & - & - & - & - & - & - & - & - & - \\ 
         PAConv\cite{2021.PAConv} &-& 73.0& 66.6& 94.6& \textbf{98.6}& 82.4& 0.0& 26.4& 58.0& 60.0& 89.7& 80.4& 74.3& 69.8& 73.5& 57.7\\
         ASSANet-L\cite{assanet} & - & - & 66.8 & - & - & - & - & - & - & - & - & - & - & - & - & -  \\ 
         Point Transformer\cite{2021.PointTransformer}& 90.8 & 76.5 & 70.4 & 94.0 & 98.5 & \textbf{86.3} & 0.0 & 38.0 & \textbf{63.4} & 74.3 & 89.1 & 82.4 & 74.3 & \textbf{80.2} & 76.0 & 59.3\\ 
         PatchFormer\cite{2022patchformer}& - & - & 68.1 & - & - & - & - & - & - & - & - & - & - & - & - & -  \\ 
         CBL \cite{CBL}& 90.6 & 75.2 & 69.4 & 93.9 & 98.4 & 84.2 & 0.0 & 37.0 & 57.7 & 71.9 & \textbf{91.7} & 81.8 & 77.8 & 75.6 & 69.1 & 62.9 \\
         RepSurf-U\cite{repsurf}& 90.2& 76.0 & 68.9 & - & - & - & - & - & - & - & - & - & - & - & - & - \\
         StratifiedFormer*\cite{2022.straitifiedformer}  & \textbf{91.5} & \textbf{78.1} & \textbf{72.0 }&\textbf{96.2}& \textbf{98.7} &\textbf{85.6}& 0.0 &\textbf{46.1} &60.0& \textbf{76.8}& \textbf{92.6}& 84.5& 77.8 &75.2& 78.1& \textbf{64.0}\\ 
         \hline
         PointNet++\cite{2017.PointNet++}& 83.0 & - & 53.5 & - & - & - & - & - & - & - & - & - & - & - & - & -\\ 
         PointNeXt-L\cite{https://2022pointnext}&90.1	&76.1	&69.5	&94.0	&98.5	&83.5	&0.0	&30.3	&57.3	&74.2	&82.1	&91.2	&74.5	&75.5	&76.7	&58.9\\
         PointNeXt-XL\cite{https://2022pointnext}&90.7	&77.5	&70.8	&94.2	&98.5	&84.4	&0.0	&37.7	&59.3	&74.0	&83.1	&91.6	&77.4	&77.2	&78.8	&60.6\\
         \textbf{PointVector-L(Ours)}&90.8&77.3&71.2&94.8&98.2&84.1&0.0&31.7&60.0&\textbf{77.7}&83.7&\textbf{91.9}&\textbf{81.8}&\textbf{78.9}&\textbf{79.9}&\textbf{63.3}\\
         \textbf{PointVector-XL(Ours)}&\textbf{91.0}&\textbf{78.1}&\textbf{72.3}&\textbf{95.1}&\textbf{98.6}&85.1&0.0&\textbf{41.4}&\textbf{60.8}&76.7&84.4&\textbf{92.1}&\textbf{82.0}&77.2&\textbf{85.1}&61.4\\
         \hline
    \end{tabular}}
    \caption{\textbf{Semantic segmentation on S3DIS Area5.} * denotes StratifiedFormer use 80k points as input points. The highest and second scores are marked in bold.}
    \label{tab:area5}
    \vspace{-0.3cm}
\end{table*}
\subsection{3D Object Classification on ScanObjectNN}
\begin{table}[htbp]
\small
    \centering
    \resizebox{\linewidth}{!}{
    \begin{tabular}{l|cc|cc}
    \hline
         Method&OA&mAcc&Params.&Throughput  \\
         &\%&\%&M&ins./sec.\\
    \hline
PointNet\cite{2017.PointNet}&68.2  &63.4 & 3.5  & \textbf{4212}  \\ 
PointCNN\cite{2018.PointCNN}&78.5  &75.1 & 0.6  & 44 \\
DGCNN\cite{2019.DGCNN}&78.1 &73.6 & 1.8   & 402  \\  
GBNet\cite{2022.GB-Net}& 80.5 & 77.8 & 8.8 & 194\\
PRANet\cite{2021.PRA-Net}&82.1  & 79.1&  2.3 & 493 \\ 
PointMLP\cite{2022.PointMLP}& $85.4\pm1.3$ & $83.9\pm1.5$ & 13.2  & 191\\ 
RepSurf-U\cite{repsurf}& 86.0 & 83.1 & 6.8 & - \\  
\hline
 PointNet++\cite{2017.PointNet++} & 77.9 &75.4 & 1.5   & 1872 \\ 
PointNeXt-S\cite{https://2022pointnext} & $\textbf{87.7}\pm0.4$  &  $\textbf{85.8}\pm0.6$ & 1.4  & 2040 \\ 
\textbf{PointVector-S(Ours)}&$\textbf{87.8}\pm0.4$&$\textbf{86.2}\pm0.5$&1.55&901\\
\hline
    \end{tabular}}
    \caption{\textbf{Object classification on ScanObjectNN.} The highest and second
scores are marked in bold.}
    \label{tab:scanobject}
    \vspace{-0.3cm}
\end{table}
ScanObjectNN\cite{2016.Scan} contains approximately 15,000 real scanned objects that are categorized into 15 classes with 2,902 unique object instances. The dataset has significant challenges due to occlusion and noise. As with PointNeXt, we chose the hardest variant PB\_T50\_RS of ScanObjectNN and report the mean$\pm$std Overall Accuracy and Mean Accuracy score. For our model in ScanObjectNN, the number of neighbors in SetAbstraction is 32. As shown in table.\ref{tab:scanobject}, our PointVector-S model achieves a comparable performance on ScanObjectNN in $OA$, while outperforms PointNeXt-S by 0.4\% in $mAcc$. This illustrates that our approach is not more biased toward certain categories and is relatively robust. Our approach is at a disadvantage in terms of speed and scale compared to the SA module. Since we introduce high-dimensional vectors, we generate more computations before the reduction compared to the standard SA module. Due to group convolution operations and trigonometric functions, there is a speed bottleneck. Although the inference speed is slower than PointNeXt, we are still faster than other methods\cite{2022.PointMLP}\cite{2019.DGCNN}\cite{2022.GB-Net}. Our method does not perform well on the classification task, where the downsampling phase of the classification task requires a max reduction function to retain salient contour information.
\subsection{3D Object Part Segmentation on ShapeNetPart}
\begin{table}[htbp]
    \centering
    \small
    \resizebox{0.8\linewidth}{!}{
    \begin{tabular}{l|cc}
\toprule
\textbf{Method}  & Ins.mIoU  & Throughput \\ 
\midrule
PointNet\cite{2017.PointNet} & 83.7& \textbf{1184}\\
DGCNN\cite{2019.DGCNN}  & 85.2 &147\\
KPConv\cite{2019.KPConv}& 86.2 & 44 \\
3D-GCN\cite{3D-GCN}&85.1&-\\
CurveNet\cite{2021.CurveNet}   & 86.8 & 97 \\
ASSANet-L\cite{assanet} & 86.1 & 640 \\
Point Transformer\cite{2021.PointTransformer} & 86.6 & 297 \\ 
PointMLP\cite{2022.PointMLP} & 86.1& 270\\
Stratifiedformer~\cite{2022.straitifiedformer} & 86.6&  398 \\ 
\midrule
PointNet++\cite{2017.PointNet++} & 85.1 & 560 \\
PointNeXt-S*\cite{https://2022pointnext} & 86.5& 776\\ 
PointNeXt-S* (C=64) & \textbf{86.9$\pm$0.1} & 330 \\
PointNeXt-S* (C=160) &\textbf{87.2}& 75\\ 
\textbf{PointVector-S(Ours)}&86.5&446\\
\textbf{PointVector-S(C=64)}&\textbf{86.9}&211\\
\bottomrule
\end{tabular}}
    \caption{\textbf{Object Part Segmentation on ShapeNetPart.} *Our evaluation results on this task alone are not consistent with the throughput results derived from that paper. Other works we did not test one by one.}
    \label{tab:shapenet}
    \vspace{-0.3cm}
\end{table}
ShapeNetPart\cite{2016.shapenet} is an object-level dataset for part segmentation. It consists of 16,880 models from 16 different shape categories, 2-6 parts for each category, and 50 part labels in total. As shown in Tab.\ref{tab:shapenet}, our PointVector-S and PointVector-S\_C64 models both achieve results that are comparable to PointNeXt. For the PointNeXt-S model with C=160, the number of parameters is large, and we do not give a corresponding version of the model.
\subsection{Ablation Study}
We perform ablation experiments at S3DIS to verify the effectiveness of the module, and because PointVector-XL is too large, we make changes to PointVector-L. To make the comparison fair, we did not change the training parameters.
\paragraph{Vector-oriented Point Set Abstraction.}
\begin{table}[htbp]
    \centering
    \small
    \begin{tabular}{l|cccc}
    \toprule
        Method &OA&mAcc&mIOU&Params  \\
        &\%&\% &M\\
        \midrule
         max+FC*&90.6&76.4&70.6&6.35\\
         Conv&90.4&75.7&69.4&24.56 \\
         GroupConv&90.6&76.5&70.8&4.76\\
         sum+FC&90.7&76.6&71.0&6.35\\
         max+GroupConv&90.6& 76.2& 70.6&4.71\\
         sum+GroupConv&90.8&77.3&71.2&4.71\\
        \bottomrule
    \end{tabular}
    \caption{Core operation of VPSA. We abstract the module into sum and GroupConv operations, and replace this part. FC means Channel-FC as Linear. * means it acts as a baseline.}
    \label{tab:VPSA}
    \vspace{-0.3cm}
\end{table}
We abstract the module into two key operations: sum and GroupConv(groups=Channel), which shows that this part of the module is channel independent, so we add a FC to mix the channel information. Considering that the channel information is already mixed using non-GroupConv operations, the channel mixing Linear will be removed. The convolution and grouped convolution parts have a convolution kernel size of $1\times k$ and a stride size of 1. As shown in Tab.\ref{tab:VPSA}, the direct use of fixed convolution brings a large number of parameters and fits very poorly with the irregular structure of the point cloud. max+FC shows better performance because intuitively aggregating features with higher dimensionality retains more information. GroupConv obtains a lower mIOU because it assigns independent weights to each element of the group; however, the three elements of a 3D vector of a channel should be given the same weight when summing. Furthermore, sum+FC is not very different from sum+GroupConv because GroupConv and channel mixing Linear can be combined into a specific layer of FC. In contrast, sum+GroupConv has the smallest number of parameters and best performance, so we chose it.
\begin{table}[htbp]
    \centering
    \small
    \begin{tabular}{l|cccc}
    \toprule
    Method&OA&mAcc&mIOU&Params\\
    &\%&\%&\%&M \\
    \midrule
        MLP &91.0 &76.5&70.8&5.55 \\
        Linear+direction&90.8&76.6&70.8&5.55\\
        Linear+rotation &90.8&77.3&71.2&4.71\\
        \bottomrule
    \end{tabular}
    \caption{Methods for obtaining vector representations. }
    \label{tab:rotation}
    \vspace{-0.3cm}
\end{table}
\paragraph{Extended Vector From Scalar.}To verify the effectiveness of our vector rotation-based method, we compare it with two other methods. As shown in Tab.\ref{tab:rotation}, MLP is represented by two Linear layers and a ReLU activation and BatchNorm layers. Linear+direction means that Linear predicts the vector modulus length, then uses MLP to obtain the unit vector as direction, and the final modulus length is multiplied by the unit vector. The rotation-based vector expansion method proposed in Section \ref{extend vector} is ahead of other methods and has fewer parameters. This shows that the rotation-based approach can use fewer parameters to obtain a vector representation more suitable for neighbor features.
\begin{table}[htbp]
    \centering
    \small
    \begin{tabular}{l|cccc}
    \toprule
         Method&OA&mAcc&mIOU&Params  \\
         &\%&\%&\%&M\\ 
         \midrule
         Scalar&90.4& 76.1&69.8&3.87\\
         2D vector&90.4&77.2&70.9&3.9\\
         3D vector&90.8&77.3&71.2&4.7\\
         \bottomrule
    \end{tabular}
    \caption{Different dimensional vector.}
    \label{tab:dimension}
    \vspace{-0.3cm}
\end{table}
\paragraph{Vector dimension.}We need to explore the connection between the effect of vector representation and dimension. Intuitively, higher dimensional vectors will be more expressive of features than lower dimensional vectors. Tab.\ref{tab:dimension} shows that the 3D vector has a better ability to express the features and that the increase in the number of parameters is not very large. The mIOU without our vector representation is still higher than the results of PointNeXt. We will discuss the validity of the other parts of our network in the supplementary material.
\paragraph{Robustness. }Table.\ref{robust} shows that our method is extremely robust to various perturbations as Stratified Transformer. The ball query we use cannot get the same neighbors in the scaled point cloud. If the query radius is scaled together, then mIOU is invariant. It indicates that our method also has scale invariance. 

\begin{table}[htbp]
\centering
\resizebox{\linewidth}{!}{
\begin{tabular}{l|c|cccccccc}
\toprule
Method            & None  & $\pi$/2   &  $\pi$    & $3 \pi$ /2  & +0.2  & -0.2  & $\times 0.8$  & $\times 1.2$  & jitter  \\
\hline
PointNet++ [25]       & 59.75 & 58.15 & 58.18 & 58.19 & 22.33 & 29.85 & 56.24 & 59.74 & 59.05   \\
PointTr [51]&70.36&65.94&67.78&65.72&70.44&70.43&65.73&66.15&59.67\\
Stratified & 71.96 & 72.59 & 72.37 & 71.86 & 71.99 & 71.93 & 70.42 & 71.21 & 72.02   \\
\hline
Ours              & 72.29 & 72.27 & 72.30 & 72.32 & 72.29 & 72.29 & 69.34 & 69.26 & 72.16  \\
\bottomrule
\end{tabular}
}\caption{Robustness study on S3DIS (mIOU $\%$). We apply z-axis rotation ($\pi$/2, $\pi$, $3 \pi$/2), shifting (±0.2), scaling (×0.8, ×1.2) and jitter in testing. PointTr: Point Transformer. Stratified: Stratified Transformer.}
\label{robust}
\vspace{-0.3cm}
\end{table}

\section{Conclusion and Limitation.}
We introduce PointVector, which achieves state-of-the-art results on the S3DIS semantic segmentation task. Our vector-oriented point set abstraction improves local feature aggregation with fewer parameters. The rotation-based vector expansion method bridges the gap between vector representation and standard feature forms. By optimizing two independent perspectives, it achieves better results. Additionally, our method exhibits robustness to various perturbations. It is noteworthy that further exploration of vector representation’s meaning may reveal additional applications, i.e. dominant neighbor selection.

The speed of our approach is constrained by the grouped convolution implementation. An interesting avenue for future work includes exploring rotations above three dimensions and decomposing four-dimensional rotations into combinations of plane rotations. Additionally, summing after component alignment aligns with our assumptions better than scalar projection.
\section*{Acknowledgement}
This work was supported in part by the National Key Research and Development Program of China under Grant 2020YFB2103803.

\appendix
\section{Preliminary}
\subsection{Problem of WaveMLP.}
WaveMLP\cite{2022WaveMLP} views the patch of each picture as a wave representation, and considers that the feature of that patch should have two attributes, phase and amplitude, with amplitude representing the actual property of the feature and phase modulating the amplitude that this wave exhibits at a moment. It thus considers that the feature extraction of the patches can be viewed as a superposition of waves. However, there is an important problem, WaveMLP gets an absolute representation of a patch, i.e. the patch is the same when participating in aggregation in any local region. The representation of a patch should be different in different local regions, so we focus on modulating the feature aggregation in local regions. That is, we use a vector representation to better express the relative relationship between neighbor points and centroids in the local region.

In addition, WaveMLP use GroupConv\cite{Groupconv.imagenet.deepcnn} to implement the aggregation and projection process with kernel sizes of $1\times 7$ and $7\times 1$. In this paper we take the form of a combination of the reduction function and GroupConv for aggregation. We give an example of why the original GroupConv is not suitable for this representation of vectors. We take two-dimensional vectors $(x_1,y_1)$ and $(x_2,y_2)$ as an example. The vectors are represented in coordinate form, and then the original vector aggregation method can be formulated as:
\begin{equation}\label{a1a3=a2a4}
\begin{aligned}
    f_{12}&=(w_1(x_1,y_1)+w_2 (x_2,y_2))\cdot(w_3,w_4)^T\\
    &=w_3 w_1 x_1+w_3 w_2 x_2+w_4 w_1 y_1+w_4 w_2 y_2\\
    &=a_1 x_1+a_2 x_2+a_3 y_1+a_4 y_2,
\end{aligned}
\end{equation}
where $f_{12}$ denotes the result of aggregating two vectors, $w_1$ and $w_2$ are the weights of two vectors in summation, $\{w_3,w_4\}$ is the projection matrix, and $a_i$ is the weight of each component. We can obtain the equation that should be satisfied between the coefficients of each component: $a_1*a_4=a_2*a_3$. That is, the final trained weights need to satisfy this equation for the weighted summation formula of the vectors to hold. However, the network does not impose this restriction on these parameters. So the original groupconv does not preserve the totality of the vector.
\subsection{Methodology Review.}
The point-based approach was first introduced by PointNet\cite{2017.PointNet}. We denote $f_i^{l+1}$ as the extracted feature of point \emph{i} after stage \emph{l+1}, $N_i$ as the neighbors of point \emph{i} and $n$ is the number of incoming points. The simplest point-set operator can be expressed as follows:
\begin{equation}\label{eq:sa module}
    f_i^{l+1}=R\{H\{[f_j^l,p_j-p_i]\}|j\in N_i\},
\end{equation}
where $R$ is the reduction function that aggregates features for point \emph{i} from its neighbors $N_i$ and $H$ means the shared MLPs.

The subsequent dynamic convolution-based network\cite{2019.KPConv}\cite{2021.PAConv} can be similarly represented as PointNet-like point set operators:
\begin{equation}\label{eq:dynamic conv}
    f_i^{l+1}=Sum\{\phi\{f_j^l,p_j-p_i\}\cdot f_j^l|j\in N_i\},
\end{equation}
where $\phi()$ means the dynamic weight generation function that generates dynamic weights for each point based on the input feature and location information. Eq.\ref{eq:dynamic conv} shows that the reduction function of dynamic convolution chooses sum and uses dynamic weights to generate a new $f_j$.

Similarly, the attention network\cite{2021.PointTransformer} can be expressed as a similar point set operator. The core operation can be formulated as follows:
\begin{equation}\label{eq:att}
    f_i^{l+1}=Sum\{att\{f_j^l,f_i^l,pos\}\cdot \sigma\{f_j^l,pos\}|j\in N_i\},
\end{equation}
where $att()$ means the attention function that generates attention weights for each point, $pos$ denotes the position information, and $\sigma()$ means the linear transform function without anisotropy. Eq.\ref{eq:att} shows that it uses the attention mechanism to update the features of each point \emph{j} and then uses sum as the reduction function.

Furthermore, template-based methods such as 3D-GCN make use of kernels with relative displacement vectors and weights. These weights are influenced by the cosine similarity between the relative displacement vector of the input features and the relative displacement vector of the kernel. The core operation can be formulated as follows:
\begin{equation}
\begin{aligned}
    &f_i^{l+1}=f_i\cdot kernel_c+\sum_{m=1}^{k}{max\{sim\{kernel_m,f_j\}|j\in N_i\}},\\
    &sim\{kernerl_m,f_j\}=cos\{dk_m,dp_j\}\cdot kernerl_m\cdot f_j,
\end{aligned}
\end{equation}
where \emph{k} means the kernel size, $N_i$ means the neighbors of point \emph{i}, $kernel_c$ means the center element of kernel, $cos\{dk_m,dp_j\}$ means Cosine similarity of \emph{m}-th kernerl element and \emph{j}-th point feature, $kernel_m$ means \emph{m}-th kernel element, $dk_m,dp_j$ means displacement vector of \emph{m}-th kernel element and \emph{j}-th point feature respectively.

We propose a unique method for generating new features $f_j$ by introducing a vector representation, where the direction of the vector guides the aggregation method.
\section{Architecture}

\subsection{Vector encoder}
\begin{figure}[htbp]
    \centering
    \includegraphics[width=\linewidth]{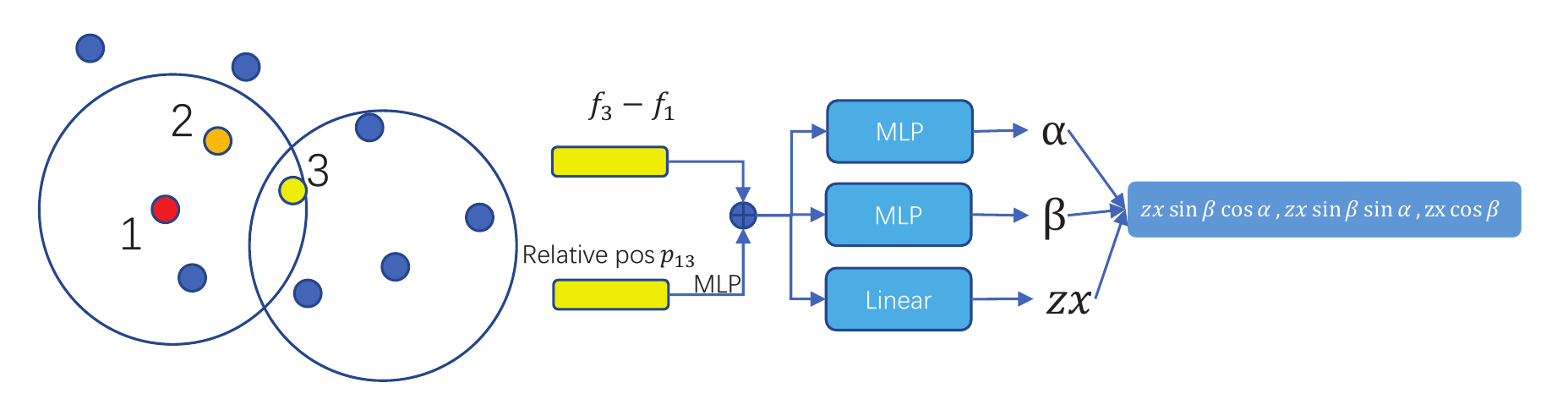}
    \caption{\textbf{The Vector encoder module.} Two angles are predicted by MLP and $zx$ is transformed by linear.}
    \label{fig:meta architecture}
\end{figure}
We provide detailed definitions in the manuscript, and we provide illustrations to illustrate the exact process. As shown in Fig.\ref{fig:meta architecture}, the local information is obtained by a combination of relative features and relative positions. Note that the sum symbol in the figure means sum and ReLU operations. We use the simplest method to predict the angles using MLP, and by default the two angles are independent of each other. For $zx$, a simple transformation is performed with linear, and then a vector representation is obtained by rotation. The vector representation $v\in R^{B,C\times3,N}$, where $B$ is the batch size, $C$ is the channel of module and $N$ is the spatial size of the input feature of the module.
\subsection{Classification architecture.}
\begin{figure}[htbp]
    \centering
    \includegraphics[width=\linewidth]{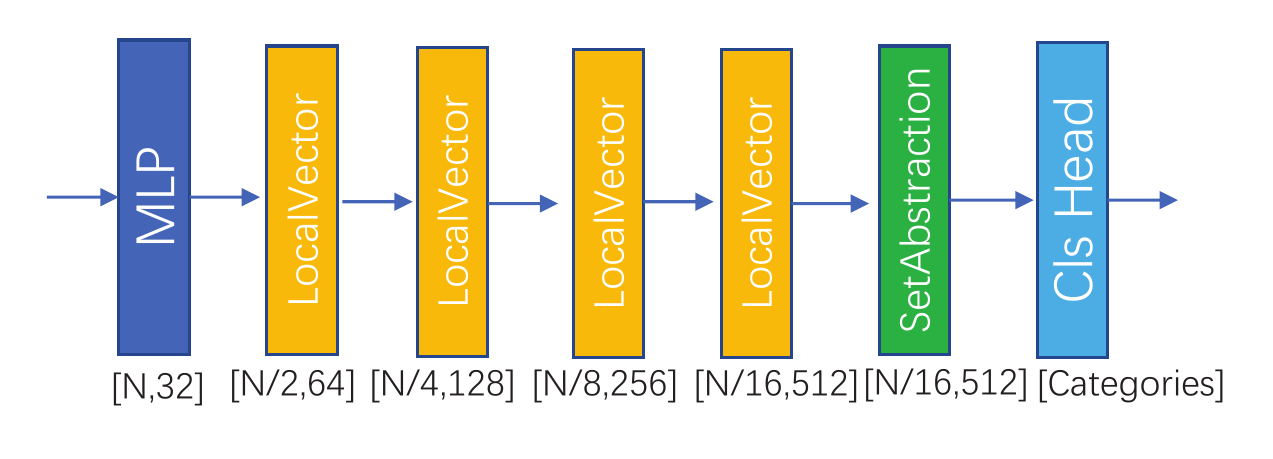}
    \caption{\textbf{The Classification architecture PointVector-S.} For comparison with PointNext\cite{https://2022pointnext}, we replaced the SetAbstract module with the LocalVector module, keeping the other parameters the same.}
    \label{fig:cls architecture}
\end{figure}
As shown in Fig.\ref{fig:cls architecture}, we use the LocalVector module to replace the 4 SetAbstract modules and keep the downsampling parameters unchanged. The last SetAbstract was originally used to aggregate all the remaining points, so we leave it as it is. In the classification task, the max reduction fuction has a greater advantage by retaining the most intense part of the variation.

\section{Experiments}
\subsection{Classification on ModelNet40}
\begin{table}[htbp]
    \centering
    \resizebox{\linewidth}{!}{
    \begin{tabular}{l|cc}
    \hline
         Method&mAcc&OA  \\
         &\%&\% \\
         \hline
         PointNet~\cite{2017.PointNet} & 86.2 & 89.2 \\ 
PointCNN\cite{2018.PointCNN} & 88.1 & 92.2  \\ 

PointConv\cite{2019.PointConv} & -- & 92.5\\ 

KPConv\cite{2019.KPConv} & - & 92.9  \\ 
DGCNN\cite{2019.DGCNN} & 90.2 & 92.9  \\ 
DeepGCN \cite{deepgcn}   & 90.9  & 93.6 \\ 
ASSANet-L \cite{assanet} & - & 92.9 \\ 
Point Cloud Transformer\cite{2021.PCT} & - & 93.2 \\ 
Point Transformer\cite{2021.PointTransformer} & 90.6 & 93.7 \\ 
CurveNet\cite{2021.CurveNet}  & - & 93.8 \\ 
PointMLP\cite{2022.PointMLP} & 90.9$\pm$0.4 & 93.7$\pm$0.2  \\ 
\hline
PointNet++&- & 91.9\\
 PointNet++(PointNext)& $89.9\pm0.8$ & $92.8\pm0.1$\\
 PointNext(C=32)&$90.8\pm 0.2$&$93.2\pm 0.1$\\
 PointNext(C=64)&$90.9\pm 0.5$&$93.7\pm 0.3$\\
 \hline
 PointVector-S(C=32)&$90.3\pm0.2$&$93.2\pm0.2$\\
 PointVector-S(C=64)&$91.0\pm0.5$&$93.5\pm 0.2$\\
 \hline
    \end{tabular}}
    \caption{Object Classification on ModelNet40.}
    \label{tab:modelnet}
\end{table}
ModelNet40\cite{2015.3dshapenet} is a commonly used dataset for object classification, which is generated by 3D graphic CAD models. It has 40 object categories, each of which contains 100 unique CAD models. Recent works\cite{2022.PointMLP}\cite{repsurf}\cite{2021.MVTN} show an increasing interest in the real-world scanned dataset ScanObejectNN\cite{2016.Scan} than this synthesized 3D dataset ModelNet40. Therefore, we choose to report the results on ScanObjectNN in the manuscript. Furthermore, we report the results of our PointVector-S model on ModelNet40. We use the same parameters as PointNext: CrossEntropy loss with label smoothing, AdamW optimizer, a learning rate of 1e-3, a weight decay of 0.05, cosine learning rate decay, and a batch size of 32 for 600 epochs, while using random scaling and translation as data augmentations. As shown in table \ref{tab:modelnet}, the relatively poor performance of our model on the ModelNet40 dataset indicates the limitation of the proposed local vector representation in aggregating global information. We used hyperparameters consistent with PointNext and a training strategy that may not be suitable for our model, which may also account for the relatively poor performance. Note that our network structure on the classification task directly takes vector feature aggregation for downsampling, but max-pooling is probably the simplest and most effective method for downsampling.
\subsection{Ablation study}
There is a slight problem with the experimental setup in the manuscript, in the 6-fold cross-validation experiment we report the PointVector-L as the standard setup mentioned in the manuscript, but in the S3DIS Area5 and ablation experiments we report the setup of PointVector-L as V=[2, 2, 4, 2]. But, the max+groupconv in the manuscript is reported as V=[2,4,2,2].

\begin{table}[htbp]
\centering
\small
\resizebox{\linewidth}{!}{
\begin{tabular}{l|l|ccc} 
\hline
Method                                                                                   & size        & OA    & mAcc  & mIOU  \\
                                                                                         &             & \%    & \%    & \%    \\ 
\hline
\multirow{2}{*}{\begin{tabular}[c]{@{}l@{}}PointVector-L\\~(max+groupconv)\end{tabular}} & V=[2,4,2,2] & 90.6  & 76.2  & 70.6  \\
                                                                                         & V=[2,2,4,2] & 90.6  & 77.1  & 71.1  \\ 
\hline
\multirow{2}{*}{\begin{tabular}[c]{@{}l@{}}PointVector-L\\~(sum+groupconv)\end{tabular}} & V=[2,4,2,2] & 90.3  & 77.21 & 70.8  \\
                                                                                         & V=[2,2,4,2] & 90.8  & 77.3  & 71.2  \\
\hline
\multirow{2}{*}{PointVector-XL}                                                          & V=[3,5,3,3] & 90.8~ & 78.3  & 72.3  \\
                                                                                         & V=[3,3,5,3] & 91.0  & 76.7  & 71.1  \\ 
\hline
\end{tabular}
}
\caption{Results for models with different number of stagess on S3DIS Area5.}
\label{tab:stages}
\end{table}
\paragraph{Number of stages. }Since the PointVector-L with max+groupconv is reported by another configuration in the manuscript, we compare the two configurations here. As the tab.\ref{tab:stages} shows, the two reduction functions, max and sum, obtain very similar results, but sum has a higher mAcc and OA. This is consistent with our assumption that better results can be obtained by simply using groupconv to process vectors of each channel independently. Small and large models do not behave consistently in terms of the number of stages. This is an interesting phenomenon, but not the main point of our statement, so it will not be discussed for now.

The following experiments are reported by default as PointVector-XL [3,5,3,3], PointVector-L [2,2,4,2] if no special instructions are given.
\begin{table}[htbp]
\centering
\small
\begin{tabular}{l|cccc} 
\hline
Method           & OA                                                                            & mAcc                                                                          & mIOU                                                                                                             & Params  \\
                 & \%                                                                            & \%                                                                            & \%                                                                                                               & M       \\ 
\hline
PointNeXt-XL     & 90.7                                                                       & 77.5                                                                          & 70.8                                                                                                            & 41.6    \\
PointVector-base & 90.9 & 77.0& 71.4 & 37.2    \\
PointVector-XL   & 90.8                                                                        & 78.3                                                                          & 72.3                  & 24.1    \\
\hline
\end{tabular}
\caption{Baseline. The same experimental configuration was used for all three models.}
\label{tab:baseline}
\end{table}
\paragraph{Baseline.}Our model has some gaps in channel variations and inputs with PointNeXt. To really evaluate whether our model has a greater advantage, we reset a baseline. We take our core operations i.e. Vector encoder and reduction+groupconv+channel mixing Linear was removed and replaced with PointNeXt's MLP+max+MLPs, where the channel of first MLP was transformed from c to 3c. The new model is named PointVector-base. The tab.\ref{tab:baseline} shows that our model has a large improvement in each metric compared to baseline. Also this shows that the other parts of our model are superior compared to the original PointNeXt.
\begin{table}[htbp]
\centering
\small
\begin{tabular}{l|l|lll} 
\hline
type                      & Method        & OA   & mAcc & mIOU  \\
                          &               & \%   & \%   & \%    \\ 
\hline
\multirow{3}{*}{feature}  & $f_j-f_i$+pos     & 90.8 & 77.3 & 71.2  \\
                          & {[}$f_j-f_i$,pos] & 88.8 & 70.5 & 64.9  \\
                          & $f_j$+pos        & 90.9 & 76.6 & 70.5  \\ 
\hline
\multirow{2}{*}{residual} & linear        & 90.8 & 77.3 & 71.2  \\
                          & identity      & 90.3 & 75.8 & 69.3  \\
\hline
\end{tabular}
\caption{Other Components. $+$ means that the two are added together and then passed through the relu layer. [,] means to directly concatenate two elements.}
\label{tab:components}
\end{table}
\paragraph{Other Components.} The manuscript mentions that other operations of our model have a larger role, so we conducted ablation experiments on PointVector-L to explore the effect of both input features and residuals on the S3DIS segmentation task. Tab.\ref{tab:components} shows that the two parts of the features are added together and then relu can better fuse their information. In addition relative features are more robust than absolute features. The key is that residual uses linear compared to identity, which is a huge improvement.
\section{Visualization}
\begin{figure}[htbp]
    \centering
    \includegraphics[width=\linewidth]{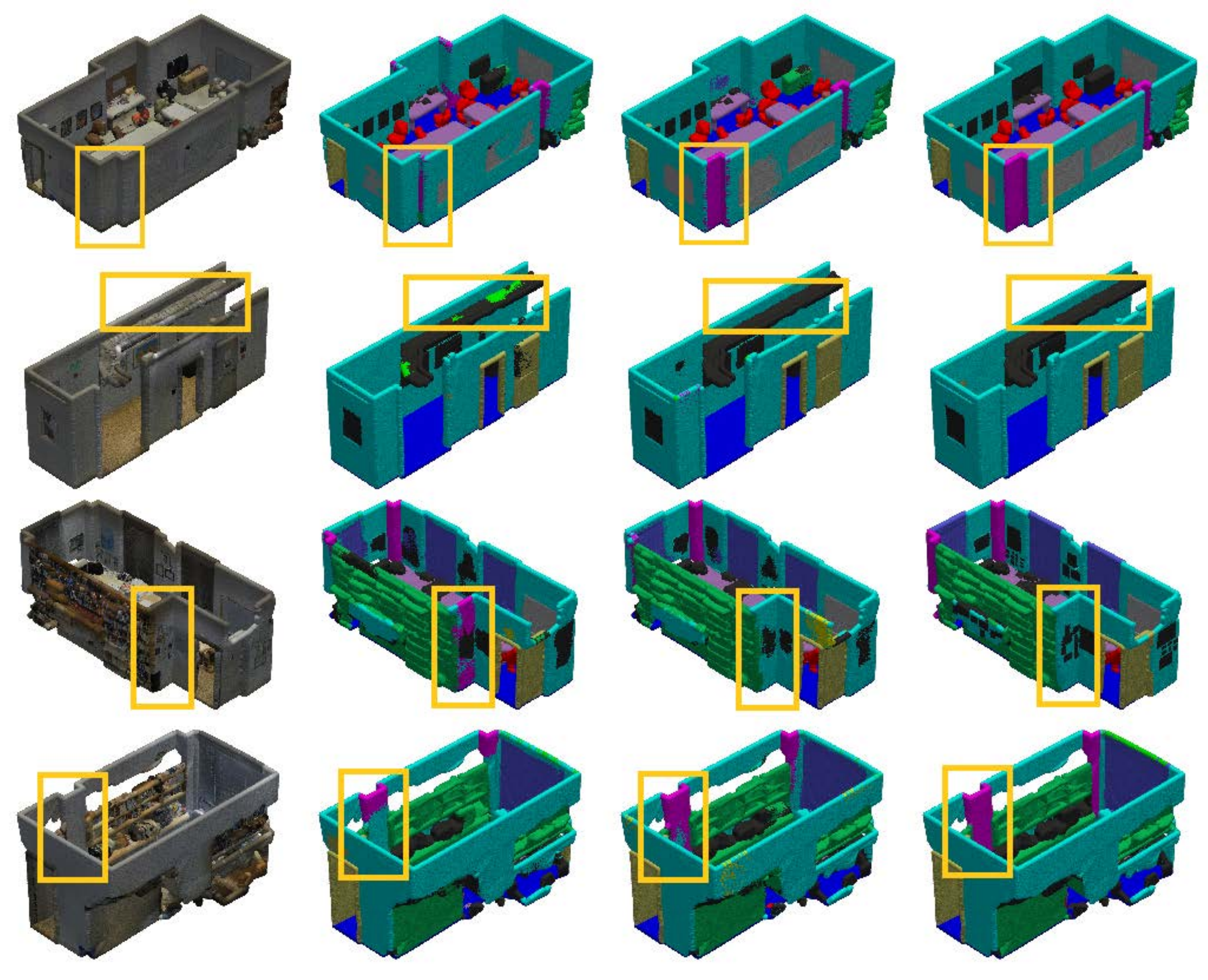}
    \caption{\textbf{Qualitative comparisons of PointNext ($2^{nd}$ column), PointVector++ ($3^{rd}$ column), and Ground Truth ($4^{th}$ column) on S3DIS semantic segmentation}. The input point cloud is visualized with original colors in the $1^{st}$ column. We have circled the different places with a paintbrush.}
    \label{fig:vis}
\end{figure}
As shown in the Fig.\ref{fig:vis}, it can be found that our model performs a little better in complex areas. This shows that we are able to extract more detail in such intensely varied areas than the max-pooling operation of PointNeXt. But we are also prone to miscalculation in flat areas, which is our disadvantage.
\section{Code release}
Since our model is based on PointNext, we used their code and added a PointVector model. Since our classification and part segmentation and semantic segmentation tasks use different model compositions, the model code is also different, and the corresponding PointVector.py needs to be replaced at runtime. We have not organized the code yet, where PATM represents the core part of our LocalVector module. In the classification and part segmentation tasks, it replaces the convs+max pooling operation in SetAbstraction. See the official instructions for PointNext for related running instructions. And on s3dis our gravity\_dim is set to 1. The code of each task is a little different, on ScanObjectNN classification task we insert leakyrelu in the two linear after the reduction function, and the relative features of the input after BN, encoder's activation function all use leakyrelu can reach 88.4\% OA, but this is not the main point of our statement, so we do not discuss it for now. The code takes time to organize and we will make it public later.

{\small
\bibliographystyle{ieee_fullname}
\bibliography{egbib}
}

\end{document}